\def\year{2019}\relax
\documentclass[letterpaper]{article} 
\usepackage{aaai19}  
\usepackage{times}  
\usepackage{helvet}  
\usepackage{courier}  
\usepackage{url}  
\usepackage{graphicx}  

\usepackage{times}
\usepackage{subfigure} 
\usepackage{algorithm}
\usepackage{algorithmic}
\usepackage{tikz}
\usetikzlibrary{matrix}

\def\E{{\rm E}}

\begin{document}
%
\title{Learning Dynamic Generator Model \\by Alternating Back-Propagation Through Time}
\author{Jianwen Xie $^{1 \ast}$, Ruiqi Gao $^{2 \ast}$, Zilong Zheng $^{2}$, Song-Chun Zhu $^{2}$, Ying Nian Wu $^{2}$\\
$^{1}$Hikvision Research Institute, Santa Clara, USA  \hspace{2mm} $^{2}$University of California, Los Angeles, USA\\
}
\maketitle
\begin{abstract}
This paper studies the dynamic generator model for spatial-temporal processes such as dynamic textures and action sequences in video data. In this model, each time frame of the video sequence is generated by a generator model, which is a non-linear transformation of a latent state vector, where the non-linear transformation is parametrized by a top-down neural network. The sequence of latent state vectors  follows a non-linear auto-regressive model, where the state vector of the next frame is a non-linear transformation of the state vector of the current frame as well as an independent noise vector that provides randomness in the transition. The non-linear transformation of this transition model can be parametrized by a feedforward neural network. We show that this model can be learned by an alternating back-propagation through time algorithm that iteratively samples the  noise vectors  and updates the parameters in the transition model and the generator model. We show that our training method can learn realistic models for dynamic textures and action patterns. 
\end{abstract}

\section{Introduction}

\subsection{The model} 

Most physical phenomena in our visual environments are spatial-temporal processes. In this paper, we study a generative model for spatial-temporal processes such as dynamic textures  and action sequences in video data.  The model is a non-linear generalization of the linear state space model proposed by  \cite{doretto2003dynamic} for dynamic textures. The model of  \cite{doretto2003dynamic} is a hidden Markov model, which consists of a transition model that governs the transition probability distribution in the state space, and an emission model that generates the observed signal by a  mapping from the state space to the signal space. In the model of  \cite{doretto2003dynamic}, the transition model is an auto-regressive model in the $d$-dimensional state space, and the emission model is a linear mapping from the $d$-dimensional state vector to the $D$-dimensional image.   In \cite{doretto2003dynamic}, the emission model is learned by treating all the frames of the input video sequence as independent observations, and the linear mapping is learned by principal component analysis via singular value decomposition. This reduces the $D$-dimensional image to a $d$-dimensional state vector. The transition model is then learned on the sequence of $d$-dimensional state vectors by a first order linear auto-regressive model. 

Given the high approximation capacity of the modern deep neural networks, it is  natural to replace the linear structures in the transition and emission models of   \cite{doretto2003dynamic} by the neural networks. This leads to the following dynamic generator model that has the following two components. (1) The emission model, which is a generator network that maps the $d$-dimensional state vector to the $D$-dimensional image via a top-down deconvolution network. (2) The transition model, where the state vector of the next frame is obtained by a non-linear transformation of the state vector of the current frame as well as an independent Gaussian white noise vector that provides randomness in the transition. The non-linear transformation can be parametrized by a feedforward neural network or multi-layer perceptron. In this model, the latent random vectors that generate the observed data are the independent Gaussian noise vectors, also called innovation vectors in   \cite{doretto2003dynamic}. The state vectors and the images can be deterministically computed from these noise vectors. 

\subsection{The learning algorithm} 

Such dynamic models have been studied in the computer vision literature recently, notably \cite{tulyakov2017mocogan}. However, the models are usually trained by the generative adversarial networks (GAN) \cite{goodfellow2014generative} with an extra discriminator network that seeks to distinguish between the observed data and the synthesized data generated by the dynamic model. Such a model may also be learned by variational auto-encoder (VAE) \cite{kingma2013auto} together with an inference model that infers the sequence of noise vectors from the sequence of observed frames. Such an inference model may require a sophisticated design. 

In this paper, we show that it is possible to learn the model on its own using an alternating back-propagation through time (ABPTT) algorithm, without recruiting a separate discriminator model or an inference model. The ABPTT algorithm iterates the following two steps. (1) Inferential back-propagation through time, which samples the sequence of noise vectors given the observed video sequence using the Langevin dynamics, where the gradient of the log posterior distribution of the noise vectors can be calculated by back-propagation through time. (2) Learning back-propagation through time, which updates the parameters of the transition model and the emission model by gradient ascent, where the gradient of the log-likelihood with respect to the model parameters can again be calculated by back-propagation through time. 

The alternating back-propagation (ABP) algorithm was originally proposed for the static generator network \cite{HanLu2016}. In this paper, we show that it can be generalized to the dynamic generator model. In our experiments, we show that we can learn the dynamic generator models using the ABPTT algorithm for dynamic textures and action sequences. 

Two advantages of the ABPTT algorithm for the dynamic generator models are convenience and efficiency. The algorithm can be easily implemented without designing an extra network. Because it only involves back-propagations through time with respect to a single model, the computation is very efficient. 

\subsection{Related work} 

The proposed learning method is related to the following themes of research. 

\textit{Dynamic textures.} The original dynamic texture model \cite{doretto2003dynamic} is linear in both the transition model and the emission model. Our work is concerned with a dynamic model with non-linear  transition and emission models. See also \cite{tesfaldet2017two} and references therein for some recent work on dynamic textures. 

\textit{Chaos modeling.} The non-linear dynamic generator model has been used to approximate chaos in a recent paper \cite{pathak2017using}. In the chaos model, the innovation vectors are given as inputs, and the model is deterministic. In contrast, in the model studied in this paper, the innovation vectors are independent Gaussian noise vectors, and the model is stochastic. 

\textit{GAN and VAE.} The dynamic generator model can also be learned by GAN or VAE.  See  \cite{tulyakov2017mocogan} \cite{saito2017temporal} and \cite{vondrick2016generating} for recent video generative models based on GAN. However,  GAN does not infer the latent noise vectors. In VAE \cite{kingma2013auto}, one needs to design an inference model for the sequence of noise vectors, which is a non-trivial task due to the complex dependency structure. Our method does not require an extra model such as a discriminator in GAN or an inference model in VAE. 

\textit{Models based on spatial-temporal filters or kernels.} The patterns in the video data can also be modeled by spatial-temporal filters by treating the data as 3D (2 spatial dimensions and 1 temporal dimension), such as a 3D energy-based model \cite{xie2017synthesizing} where the energy function is parametrized by a 3D bottom-up ConvNet, or a 3D generator model \cite{HanLu2019} where a top-down 3D ConvNet maps a latent random vector to the observed video data. Such models do not have a dynamic structure defined by a transition model, and they are not convenient for predicting future frames. 

\subsection{Contribution} 
The main contribution of this paper lies in the combination of the dynamic generator model and the alternating back-propagation through time algorithm. Both the model and algorithm are simple and natural, and their combination can be very useful for modeling and analyzing spatial-temporal processes. The model is  one-piece in the sense that (1) the transition model and emission model are integrated into a single latent variable model. (2) The learning of the dynamic model is end-to-end, which is different from \cite{HanLu2016}'s treatment. 
(3) The learning of our model does not need to recruit a discriminative network (like GAN) or an inference network (like VAE), which makes our method simple and efficient in terms of computational cost and model parameter size.

 \section{Model and learning algorithm}

\subsection{Dynamic generator model} 

Let $X = (x_t, t = 1, ..., T)$ be the observed video sequence, where $x_t$ is a frame at time $t$. The dynamic generator model consists of the following two components: 
 \begin{eqnarray}
 && s_{t} = F_\alpha(s_{t-1}, \xi_t),  \label{eq:t} \\
 && x_t = G_\beta(s_t) + \epsilon_t, \label{eq:e}
 \end{eqnarray}
where $t = 1, ..., T$.  (\ref{eq:t}) is the transition model, and (\ref{eq:e}) is the emission model. $s_t$ is the $d$-dimensional hidden state vector. $\xi_t \sim {\rm N}(0, I)$ is the noise vector of a certain dimensionality. The Gaussian noise vectors $(\xi_t, t = 1, ..., T)$ are independent of each other. The sequence of $(s_t, t = 1, ..., T)$ follows a non-linear auto-regressive model, where the noise vector $\xi_t$ encodes the randomness in the transition from $s_{t-1}$ to $s_{t}$ in the $d$-dimensional state space.  $F_\alpha$ is a feedforward neural network or multi-layer perceptron, where $\alpha$ denotes the weight and bias parameters of the network. We can adopt a residual form \cite{he2016deep} for $F_\alpha$ to model the change of the state vector. $x_t$ is the $D$-dimensional image, which is generated by the $d$-dimensional hidden state vector $s_t$.  $G_\beta$ is a top-down convolutional network (sometimes also called deconvolution network),  where $\beta$ denotes the weight and bias parameters of this top-down network. $\epsilon_t \sim {\rm N}(0, \sigma^2 I_D)$ is the residual error.  We let $\theta = (\alpha, \beta)$  denote all the model parameters. 

Let $\xi = (\xi_t, t = 1, ..., T)$. $\xi$ consists of the latent random vectors that need to be inferred from $X$.  Although $x_t$ is generated by the state vector $s_t$, $S = (s_t, t = 1, ..., T)$ are generated by $\xi$.  In fact, we can write $X = H_\theta(\xi) + \epsilon$, where $H_\theta$ composes $F_\alpha$ and $G_\beta$ over time, and $\epsilon = (\epsilon_t, t = 1, ..., T)$ denotes the observation errors. 

\subsection{Learning and inference algorithm} 

Let $p(\xi)$ be the prior distribution of $\xi$. Let $p_\theta(X|\xi) \sim {\rm N}(H_\theta(\xi), \sigma^2 I)$ be the conditional distribution of $X$ given $\xi$, where $I$ is the identity matrix whose dimension matches that of $X$. The marginal distribution is $p_\theta(X) = \int p(\xi) p_\theta(X|\xi) d\xi$ with the latent variable $\xi$ integrated out. We estimate the model parameter $\theta$ by the maximum likelihood method that maximizes the observed-data log-likelihood $\log p_\theta(X)$, which is analytically intractable. In contrast, the complete-data log-likelihood $\log p_\theta(\xi, X)$, where $p_\theta(\xi, X) = p(\xi) p_\theta(X|\xi)$, is analytically tractable. The following identity links the gradient of the observed-data log-likelihood $\log p_\theta(X)$ to the gradient of the complete-data log-likelihood $\log p_\theta(\xi, X)$:

\begin{eqnarray} 
   \frac{\partial}{\partial \theta} \log p_\theta(X)&=&\frac{1}{p_\theta(X)} \frac{\partial}{\partial \theta} p_\theta(X) \nonumber\\ 
   &=& \frac{1}{p_\theta(X)} \int \left[ \frac{\partial}{\partial \theta} \log p_\theta(\xi, X)  \right] p_\theta(\xi, X) d\xi \nonumber \\
   &=& \E_{p_\theta(\xi|X)} \left[  \frac{\partial}{\partial \theta} \log p_\theta(\xi, X)  \right],  \label{eq:gr}
\end{eqnarray}
where $p_\theta(\xi|X) = p_\theta(\xi, X)/p_\theta(X)$ is the posterior distribution of the latent $\xi$ given the observed $X$. The above expectation can be approximated by Monte Carlo average. Specifically, we sample from the posterior distribution $p_\theta(\xi|X)$ using the Langevin dynamics: 
\begin{eqnarray} 
    \xi^{(\tau+1)} = \xi^{(\tau)}   + \frac{\delta^2}{2} \frac{\partial}{\partial \xi} \log p_\theta(\xi^{(\tau)}|X) + \delta z_\tau, \label{eq:Langevin}
\end{eqnarray}
where $\tau$ indexes the time step of the Langevin dynamics (not to be confused with the time step of the dynamics model,  $t$), $z_\tau \sim {\rm N}(0, I)$ where $I$ is the identity matrix whose dimension matches that of $\xi$, and $\xi^{(\tau)} = (\xi^{(\tau)}_t, t = 1, ..., T)$ denotes all the sampled latent noise vectors at time step $\tau$. $\delta$ is the step size of the Langevin dynamics. We can correct for the finite step size by adding a Metropolis-Hastings acceptance-rejection step. After sampling $\xi \sim p_\theta(\xi|X)$ using the Langevin dynamics, we can  update $\theta$ by stochastic gradient ascent 
\begin{eqnarray} 
     \Delta \theta \propto   \frac{\partial}{\partial \theta} \log p_\theta(\xi, X),  \label{eq:learn}
\end{eqnarray}
where the stochasticity of the gradient ascent comes from the fact that we use Monte Carlo to approximate the expectation in  (\ref{eq:gr}). The learning algorithm iterates the following two steps. (1) Inference step: Given the current $\theta$, sample $\xi$ from $p_\theta(\xi|X)$ according to (\ref{eq:Langevin}). (2) Learning step: Given $\xi$, update $\theta$ according to (\ref{eq:learn}). We can use a warm start scheme for sampling in step (1). Specifically, when running the Langevin dynamics, we start from the current $\xi$, and run a finite number of steps. Then we update $\theta$ in step (2) using the sampled $\xi$. Such a stochastic gradient ascent algorithm has been analyzed by \cite{younes1999convergence}. 

Since $\frac{\partial}{\partial \xi} \log p_\theta(\xi|X) = \frac{\partial}{\partial \xi} \log p_\theta(\xi, X)$, both steps (1) and (2) involve derivatives of 
\begin{eqnarray}
    \log p_\theta(\xi, X) = -\frac{1}{2} \left[ \|\xi\|^2 + \frac{1}{\sigma^2} \|X - H_\theta(\xi)\|^2\right] + {\rm const}, \nonumber
 \end{eqnarray}
where the constant term does not depend on $\xi$ or $\theta$. Step (1) needs to compute the derivative of $\log p_\theta(\xi, X)$ with respect to $\xi$. Step (2) needs to compute the derivative of $\log p_\theta(\xi, X)$ with respect to $\theta$. Both can be computed by back-propagation through time. Therefore the algorithm is an alternating back-propagation through time algorithm. Step (1) can be called inferential back-propagation through time. Step (2) can be called learning back-propagation through time. 

To be more specific,  the complete-data log-likelihood $\log p_\theta(\xi, X)$ can be written as (up to an additive constant, assuming $\sigma^2 = 1$)
 \begin{eqnarray}
    L(\theta, \xi) =  -\frac{1}{2} \sum_{t=1}^{T} \left[ \|x_t - G_\beta(s_t)\|^2 +  \|\xi_t\|^2\right]. 
 \end{eqnarray}
The derivative with respect to $\beta$ is 
\begin{eqnarray}
  \frac{\partial L}{\partial \beta}  =   \sum_{t=1}^{T} (x_t - G_\beta(s_t)) \frac{\partial G_\beta(s_t)}{\partial \beta}. \label{eq:beta}
 \end{eqnarray}
 The derivative with respect to $\alpha$ is 
 \begin{eqnarray}
    \frac{\partial L}{\partial \alpha}  =   \sum_{t=1}^{T} (x_t - G_\beta(s_t)) \frac{\partial G_\beta(s_t)}{\partial s_t}  \frac{\partial s_t}{\partial \alpha}, \label{eq:alpha}
 \end{eqnarray}
 where $ \frac{\partial s_t}{\partial \alpha}$ can be computed recursively.
  To infer $\xi$, for any fixed time point $t_0$, 
 \begin{eqnarray}
 \frac{\partial L}{\partial \xi_{t_0}} =  \sum_{t=t_0+1}^{T} (x_t - G_\beta(s_t)) \frac{\partial G_\beta(s_t)}{\partial s_t}  \frac{\partial s_t}{\partial \xi_{t_0}} - \xi_{t_0}, 
 \end{eqnarray}
  where $ \frac{\partial s_t}{\partial \xi_{t_0}}$ can again be computed recursively.
  
 A  minor issue is the initialization of the transition model. We may assume that $s_0 \sim {\rm N}(0, I)$. In the inference step, we can sample $s_0$ together with $\xi$ using the Langevin dynamics. 
 
It is worth mentioning the difference between our algorithm and the variational inference. While variational inference is convenient for learning a regular generator network, for the dynamic generator model studied in this paper, it is not a simple task to design an inference model that infers the sequence of latent vectors $\xi = (\xi_t, t = 1, ..., T)$ from the sequence of $X = (x_t, t = 1, ..., T)$. In contrast, our learning method does not require such an inference model and can be easily implemented. The inference step in our model can be done via directly sampling from the posterior distribution $p_\theta(\xi|X)$, which is powered by back-propagation through time. Additionally, our model directly targets maximum likelihood, while model learning via variational inference is to maximize a lower bound.  
  
  \subsection{Learning from multiple sequences} 
  
 We can learn the model from multiple sequences of different appearances but of similar motion patterns. Let $X^{(i)} = (x^{(i)}_t, t = 1, ..., T)$ be the $i$-th training sequence, $i  = 1, ..., n$. We can use an appearance (or content) vector $a^{(i)}$ for each sequence to account for the variation in appearance. The model is of the following form 
 \begin{eqnarray}
 && s^{(i)}_{t} = F_\alpha(s^{(i)}_{t-1}, \xi^{(i)}_t),   \label{eq:t1} \\
 && x^{(i)}_t = G_\beta(s^{(i)}_t, a^{(i)}) + \epsilon^{(i)}_t,   \label{eq:e1}
 \end{eqnarray}
 where $a^{(i)} \sim {\rm N}(0, I)$, and $a^{(i)}$ is fixed over time for each sequence $i$.  To learn from such training data, we only need to add the Langevin sampling of $a^{(i)}$. If the motion sequences are of different motion patterns, we can also introduce another vector $m^{(i)} \sim {\rm N}(0, I)$ to account for the variations of motion patterns, so that the transition model becomes $s^{(i)}_{t} = F_\alpha(s^{(i)}_{t-1}, \xi^{(i)}_t, m^{(i)})$ with $m^{(i)}$ fixed for the sequence $i$. 
 
Recently \cite{tulyakov2017mocogan} studies a similar model where the transition model is modeled by a recurrent neural network (RNN) with another layer of hidden vectors. \cite{tulyakov2017mocogan}  learns the model using GAN. In comparison, we use a simpler Markov transition model and we learn the model by alternating back-propagation through time. Even though the latent state vectors follow a Markovian model, the observed sequence is non-Markovian. 

\begin{algorithm}
\caption{Learning and inference by alternating back-propagation through time (ABPTT)}
\label{code:ccl}
\begin{algorithmic}[1]

\REQUIRE

(1) training sequences $\{X^{(i)}= (x^{(i)}_t, t = 1, ..., T), i = 1, ..., n\}$\\
(2) number of Langevin steps $l$\\
(3) number of learning iterations $N$.

\ENSURE
(1) learned parameters $\theta = (\alpha, \beta)$\\
(2) inferred noise vectors $\xi^{(i)} = (\xi^{(i)}_t, t = 1, ..., T)$.  

\item[]
\STATE Initialize $\theta = (\alpha, \beta)$. Initialize $\xi^{(i)}$ and $a^{(i)}$. Initialize $k = 0$. 
\REPEAT 
\STATE \textbf{Inferential back-propagation through time}: For $i = 1, ..., n$, sample $\xi^{(i)}$ and $a^{(i)}$ by running $l$ steps of Langevin dynamics according to (\ref{eq:Langevin}), starting from their current values. 
\STATE \textbf{Learning back-propagation through time}: Update $\alpha$ and $\beta$ by gradient ascent according to (\ref{eq:alpha}) and (\ref{eq:beta}).  
\STATE Let $k \leftarrow k+1$
\UNTIL $k= N$
\end{algorithmic}
\end{algorithm}

Algorithm 1 summarizes the learning and inference algorithm for multiple sequences with appearance vectors. If we learn from a single sequence such as dynamic texture, we can remove the appearance vector $a^{(i)}$, or simply fix it to a zero vector.

\section{Related models} 

In this section, we shall review related models of spatial-temporal processes in order to put our work into the big picture. 

\subsection{Two related spatial-temporal models} 

Let $X = (x_t, t = 1, ..., T)$ be the observed sequence. We have studied the following energy-based model \cite{xie2017synthesizing}: 
\begin{eqnarray}
  p(X; \theta) = \frac{1}{Z(\theta)} \exp \left[ f_\theta(X) \right],  \label{eq:en}
\end{eqnarray}
where $f_\theta(X)$ is a function of the whole sequence $X$,  which  can be defined by a bottom-up network that consists of multiple layers of spatial-temporal filters that capture the spatial-temporal patterns in $X$ at multiple layers.  $\theta$ collects all the weight and bias parameters of the bottom-up network.  The model can be learned by maximum likelihood, and the learning algorithm follows an ``analysis by synthesis'' scheme. The algorithm iterates (1) Synthesis: generating synthesized sequences from the current model by Langevin dynamics. (2) Analysis: updating $\theta$ based on the difference between the observed sequences and synthesized sequences. The two steps play an adversarial game with $f_\theta$ serving as a critic. The synthesis step seeks to modify the synthesized examples to  increase $f_\theta$ scores of the synthesized examples, while the analysis step seeks to modify $\theta$ to increase the $f_\theta$ scores of the observed examples relative to the synthesized examples. 

We have also studied the following generator model \cite{HanLu2019}
\begin{eqnarray} 
    s \sim {\rm N}(0, I_d), \; X = g_\theta(s) + \epsilon, \label{eq:ge}
\end{eqnarray}
where the latent state vector $s$ is defined for the whole sequence, and is assumed to follow a prior distribution which is $d$-dimensional Gaussian white noise. The whole sequence is then generated by a function $g_\theta(s)$ that can be defined by a top-down network that consists of multiple layers of spatial-temporal kernels. $\theta$ collects all the weight and bias parameters of the top-down network. $\epsilon$ is the Gaussian noise image sequence. This generator model can be learned by maximum likelihood, and the learning algorithm follows the alternating back-propagation method of \cite{HanLu2016}.  

In \cite{coopnets2018}, we show that we can learn the above two models simultaneously using a cooperative learning scheme. We can also cooperatively train the energy-based model (\ref{eq:en}) and the dynamic generator model (\ref{eq:t}) and (\ref{eq:e}) studied in this paper simultaneously, where the dynamic generator model serves as an approximate sampler of the energy-based model. 

Unlike the dynamic generator model (\ref{eq:t}) and (\ref{eq:e}) studied in this paper, the above two models (\ref{eq:en}) and (\ref{eq:ge}) are not of a dynamic or causal nature in that they do not directly evolve or unfold over time. 

\subsection{Action, control, policy, and cost} 

If we observe the sequence of actions ${\bf a} = (a_t, t = 1, ..., T)$ applied to the system, we can extend the forward dynamic model (\ref{eq:t}) to 
\begin{eqnarray}
s_{t} = F_\alpha(s_{t-1}, a_t, \xi_t).  \label{eq:ta}
\end{eqnarray}
The model can still be learned by alternating back-propagation through time. With a properly defined cost function, we can optimize the sequence ${\bf a} = (a_t, t = 1, ..., T)$ for control. We may also learn a policy $\pi(a_t \mid s_{t-1})$ directly from demonstrations by expert controllers. We may call the resulting model that consists of both dynamics and control policy as the controlled dynamic generator model. 

We can also learn the cost function from expert demonstrations by inverse reinforcement learning \cite{ziebart2008maximum} \cite{abbeel2004apprenticeship}, where we can generalize the above energy-based model  (\ref{eq:en})  to $p_\theta(X, {\bf a}) = \frac{1}{Z(\theta)} \exp[f_\theta(X, {\bf a})]$, where $-f_\theta(X, {\bf a})$ can be interpreted as the total cost. We can learn both the cost function and the policy cooperatively as in \cite{coopnets2018}, where we fit the energy-based model to the demonstration data while using the controlled dynamic generator model as an approximate sampler of the energy-based model. 

\subsection{Velocity field, optical flow, and physics} 

In our work, the training data are image frames of video sequences. If we are given the velocity fields over time, we can also learn the dynamic generator model from such data, such as turbulence. Even with raw image sequences, it may still be desirable to learn a model that generates the velocity fields or optical flows over time, which in turn generate the image frames over time. This will lead to a more physically meaningful model of motion, which can be considered the mental physics. In our recent work on deformable generator network \cite{xing2018deformable}, we model the deformations explicitly. We can combine the deformable generator model and the dynamic generator model. We may also consider restricting the transition model to be linear to make the state vector close to real physical variables.

\section{Experiments}


\subsection{Experiment 1: Learn to generate dynamic textures}
\label{sec:dynamicTexture}
We first learn the model for dynamic textures, which are sequences of images of moving scenes that exhibit stationarity in time. We learn a separate model from each example. The video clips for training are collected from DynTex++ dataset of \cite{ghanem2010maximum} and the Internet. Each observed video clip is prepared to be of the size 64 pixels $\times$ 64 pixels $\times$ 60 frames. We implement our model and learning algorithm in Python with Tensorflow \cite{tensorflow2015-whitepaper}. The transition model is a feedforward neural network with three layers. The network takes a 100-dimensional state vector $s_{t-1}$ and a 100-dimensional noise vector $\xi_t$ as input and produces a 100-dimensional vector $r_t$, so that $s_{t} = \tanh(s_{t-1} + r_t)$.  The numbers of nodes in the three layers of the feedforward neural network are $\{20, 20, 100\}$. The emission model is a top-down deconvolution neural network or generator model that maps the 100-dimensional state vector (i.e., $1 \times 1 \times 100$) to the image frame of size $64 \times 64 \times 3$ by 6 layers of deconvolutions with kernel size of 4 and up-sampling factor of 2 from top to bottom. The numbers of channels at different layers of the generator are $\{512, 512, 256, 128, 64, 3\}$. Batch normalization \cite{ioffe2015batch} and ReLU layers are added between deconvolution layers, and tanh activation function is used at the bottom layer to make the output signals fall within $[-1, 1]$. We use the Adam \cite{kingma2015adam} for optimization with $\beta_1= 0.5$ and the learning rate is 0.002. We set the Langevin step size to be $\delta=0.03$ for all latent variables, and the standard deviation of residual error $\sigma=1$. We run $l=15$ steps of Langevin dynamics for inference of the latent noise vectors within each learning iteration. 

Once the model is learned, we can synthesize dynamic textures from the learned model
by firstly randomly initializing the initial hidden state $s_0$, and then following Equation (\ref{eq:t}) and (\ref{eq:e}) to generate a sequence of images with a sequence of innovation vectors $\{\xi_t\}$ sampled from Gaussian distribution. In practice, we use "burn-in" to throw away some iterations at the beginning of the dynamic process to ensure the transition model enters the high probability region (i.e., the state sequence $\{s_t\}$ converges to stationarity), no matter where $s_0$ starts from.

\begin{figure}
\centering	
\hspace{0.5mm}\rotatebox{90}{\hspace{4mm}{\footnotesize obs }}	
\includegraphics[height=.14\linewidth, width=.14\linewidth]{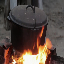}
\includegraphics[height=.14\linewidth, width=.14\linewidth]{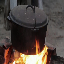}
\includegraphics[height=.14\linewidth, width=.14\linewidth]{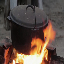}
\includegraphics[height=.14\linewidth, width=.14\linewidth]{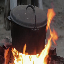}
\includegraphics[height=.14\linewidth, width=.14\linewidth]{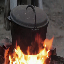}
\includegraphics[height=.14\linewidth, width=.14\linewidth]{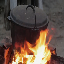}\\
\vspace{0.5mm}
\rotatebox{90}{\hspace{4mm}{\footnotesize syn1}}
\includegraphics[height=.14\linewidth, width=.14\linewidth]{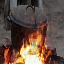}
\includegraphics[height=.14\linewidth, width=.14\linewidth]{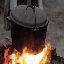}
\includegraphics[height=.14\linewidth, width=.14\linewidth]{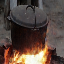}
\includegraphics[height=.14\linewidth, width=.14\linewidth]{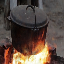}
\includegraphics[height=.14\linewidth, width=.14\linewidth]{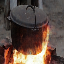}
\includegraphics[height=.14\linewidth, width=.14\linewidth]{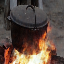}\\
\vspace{0.5mm}
\rotatebox{90}{\hspace{4mm}{\footnotesize syn2}}
\includegraphics[height=.14\linewidth, width=.14\linewidth]{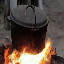} 
\includegraphics[height=.14\linewidth, width=.14\linewidth]{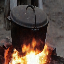} 
\includegraphics[height=.14\linewidth, width=.14\linewidth]{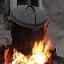}
\includegraphics[height=.14\linewidth, width=.14\linewidth]{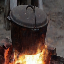}
\includegraphics[height=.14\linewidth, width=.14\linewidth]{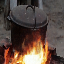}
\includegraphics[height=.14\linewidth, width=.14\linewidth]{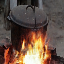}\\
(a) burning fire heating a pot\\
\vspace{0.5mm}
\hspace{0.5mm}\rotatebox{90}{\hspace{4mm}{\footnotesize obs }}	
\includegraphics[height=.14\linewidth, width=.14\linewidth]{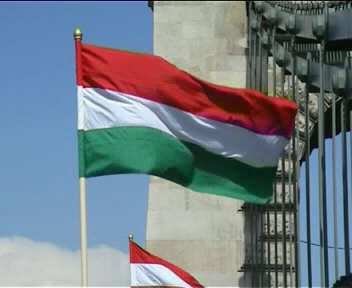}
\includegraphics[height=.14\linewidth, width=.14\linewidth]{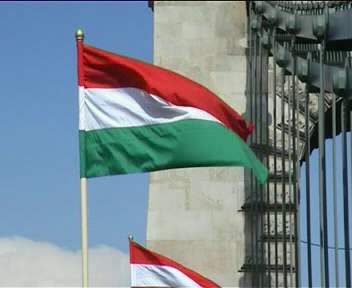}
\includegraphics[height=.14\linewidth, width=.14\linewidth]{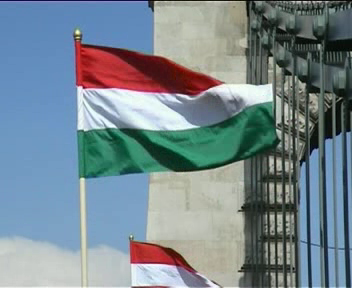}
\includegraphics[height=.14\linewidth, width=.14\linewidth]{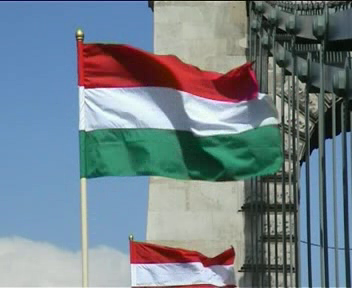}
\includegraphics[height=.14\linewidth, width=.14\linewidth]{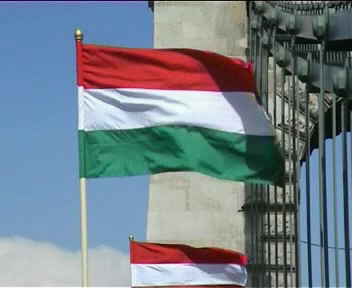}
\includegraphics[height=.14\linewidth, width=.14\linewidth]{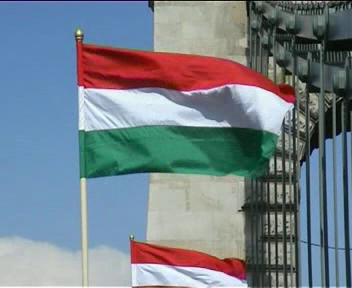}\\
\vspace{0.5mm}
\rotatebox{90}{\hspace{4mm}{\footnotesize syn1}}
\includegraphics[height=.14\linewidth, width=.14\linewidth]{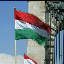}  
\includegraphics[height=.14\linewidth, width=.14\linewidth]{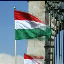}
\includegraphics[height=.14\linewidth, width=.14\linewidth]{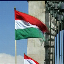}
\includegraphics[height=.14\linewidth, width=.14\linewidth]{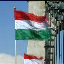}
\includegraphics[height=.14\linewidth, width=.14\linewidth]{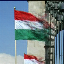}
\includegraphics[height=.14\linewidth, width=.14\linewidth]{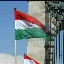}\\
\vspace{0.5mm}
\rotatebox{90}{\hspace{4mm}{\footnotesize syn2}}
\includegraphics[height=.14\linewidth, width=.14\linewidth]{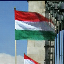}
\includegraphics[height=.14\linewidth, width=.14\linewidth]{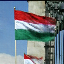}
\includegraphics[height=.14\linewidth, width=.14\linewidth]{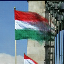}
\includegraphics[height=.14\linewidth, width=.14\linewidth]{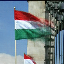}
\includegraphics[height=.14\linewidth, width=.14\linewidth]{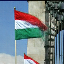}
\includegraphics[height=.14\linewidth, width=.14\linewidth]{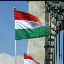}\\
(b) flapping flag\\
\vspace{0.5mm}
\hspace{0.5mm}\rotatebox{90}{\hspace{4mm}{\footnotesize obs }}	
\includegraphics[height=.14\linewidth, width=.14\linewidth]{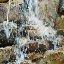}
\includegraphics[height=.14\linewidth, width=.14\linewidth]{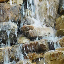}
\includegraphics[height=.14\linewidth, width=.14\linewidth]{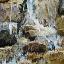}
\includegraphics[height=.14\linewidth, width=.14\linewidth]{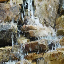}
\includegraphics[height=.14\linewidth, width=.14\linewidth]{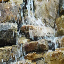}
\includegraphics[height=.14\linewidth, width=.14\linewidth]{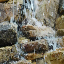}\\
\vspace{0.5mm}
\rotatebox{90}{\hspace{4mm}{\footnotesize syn1}}
\includegraphics[height=.14\linewidth, width=.14\linewidth]{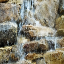}
\includegraphics[height=.14\linewidth, width=.14\linewidth]{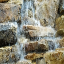}
\includegraphics[height=.14\linewidth, width=.14\linewidth]{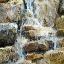}
\includegraphics[height=.14\linewidth, width=.14\linewidth]{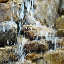}
\includegraphics[height=.14\linewidth, width=.14\linewidth]{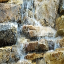}
\includegraphics[height=.14\linewidth, width=.14\linewidth]{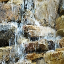}\\ 
\vspace{0.5mm}
\rotatebox{90}{\hspace{4mm}{\footnotesize syn2}}
\includegraphics[height=.14\linewidth, width=.14\linewidth]{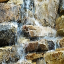}
\includegraphics[height=.14\linewidth, width=.14\linewidth]{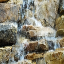}
\includegraphics[height=.14\linewidth, width=.14\linewidth]{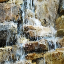}
\includegraphics[height=.14\linewidth, width=.14\linewidth]{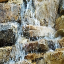}
\includegraphics[height=.14\linewidth, width=.14\linewidth]{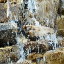}
\includegraphics[height=.14\linewidth, width=.14\linewidth]{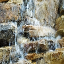}\\ 
\hspace{5mm} 
(c) waterfall\\
\vspace{0.5mm}
\hspace{0.5mm}\rotatebox{90}{\hspace{4mm}{\footnotesize obs }}	
\includegraphics[height=.14\linewidth, width=.14\linewidth]{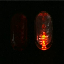}
\includegraphics[height=.14\linewidth, width=.14\linewidth]{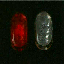}
\includegraphics[height=.14\linewidth, width=.14\linewidth]{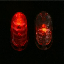}
\includegraphics[height=.14\linewidth, width=.14\linewidth]{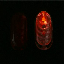}
\includegraphics[height=.14\linewidth, width=.14\linewidth]{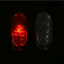}
\includegraphics[height=.14\linewidth, width=.14\linewidth]{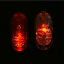}\\
\vspace{0.5mm}
\rotatebox{90}{\hspace{4mm}{\footnotesize syn1}}
\includegraphics[height=.14\linewidth, width=.14\linewidth]{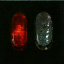}
\includegraphics[height=.14\linewidth, width=.14\linewidth]{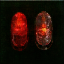}
\includegraphics[height=.14\linewidth, width=.14\linewidth]{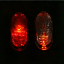}
\includegraphics[height=.14\linewidth, width=.14\linewidth]{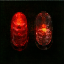}
\includegraphics[height=.14\linewidth, width=.14\linewidth]{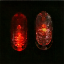}
\includegraphics[height=.14\linewidth, width=.14\linewidth]{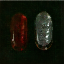}\\ 
\vspace{0.5mm}
\rotatebox{90}{\hspace{4mm}{\footnotesize syn2}}
\includegraphics[height=.14\linewidth, width=.14\linewidth]{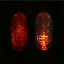}
\includegraphics[height=.14\linewidth, width=.14\linewidth]{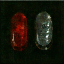}
\includegraphics[height=.14\linewidth, width=.14\linewidth]{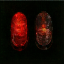}
\includegraphics[height=.14\linewidth, width=.14\linewidth]{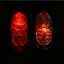}
\includegraphics[height=.14\linewidth, width=.14\linewidth]{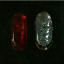}
\includegraphics[height=.14\linewidth, width=.14\linewidth]{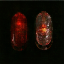}\\ 
\hspace{5mm} 
(d) flashing lights\\
\vspace{0.5mm}
\hspace{0.5mm}\rotatebox{90}{\hspace{4mm}{\footnotesize obs }}	
\includegraphics[height=.14\linewidth, width=.14\linewidth]{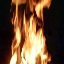}
\includegraphics[height=.14\linewidth, width=.14\linewidth]{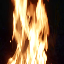}
\includegraphics[height=.14\linewidth, width=.14\linewidth]{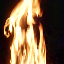}
\includegraphics[height=.14\linewidth, width=.14\linewidth]{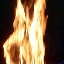}
\includegraphics[height=.14\linewidth, width=.14\linewidth]{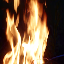}
\includegraphics[height=.14\linewidth, width=.14\linewidth]{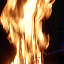}
\\
\vspace{0.5mm}
\rotatebox{90}{\hspace{4mm}{\footnotesize syn1}}
\includegraphics[height=.14\linewidth, width=.14\linewidth]{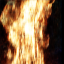}
\includegraphics[height=.14\linewidth, width=.14\linewidth]{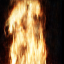}
\includegraphics[height=.14\linewidth, width=.14\linewidth]{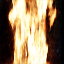}
\includegraphics[height=.14\linewidth, width=.14\linewidth]{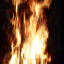}
\includegraphics[height=.14\linewidth, width=.14\linewidth]{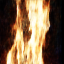}
\includegraphics[height=.14\linewidth, width=.14\linewidth]{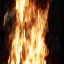}\\ 
\vspace{0.5mm}
\rotatebox{90}{\hspace{4mm}{\footnotesize syn2}}
\includegraphics[height=.14\linewidth, width=.14\linewidth]{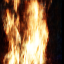}
\includegraphics[height=.14\linewidth, width=.14\linewidth]{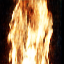}
\includegraphics[height=.14\linewidth, width=.14\linewidth]{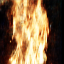}
\includegraphics[height=.14\linewidth, width=.14\linewidth]{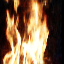}
\includegraphics[height=.14\linewidth, width=.14\linewidth]{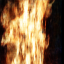}
\includegraphics[height=.14\linewidth, width=.14\linewidth]{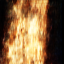}\\ 
\hspace{5mm} 
(e) flame\\ \vspace{-1mm}
\caption{Generating dynamic textures. For each category, the first row displays 6 frames of the observed sequence, and the second and third rows show the corresponding frames of two synthesized sequences generated by the learned model.}	
\label{fig:dynamicTexture}
\end{figure}

To speed up the training process and relieve the burden of computer memory, we can use truncated back-propagation through time in training our model. That is, we divide the whole training sequence into different non-overlapped chunks, and run forward and backward passes through chunks of the sequence instead of the whole sequence. We carry hidden states $\{s_t\}$ forward in time forever, but only back-propagate for the length (the number of image frames) of chunk. In this experiment, the length of chunk is set to be 30 image frames.

An ``infinite length'' dynamic texture can be synthesized from a typically ``short'' input sequence by just drawing ``infinite'' IID samples from Gaussian distribution. Figure \ref{fig:dynamicTexture} shows five results. For each example, the first row displays 6 frames of the observed 60-frame sequence, while the second and third rows display 6 frames of two synthesized sequences of 120 frames in length, which are generated by the learned model. 

Similar to \cite{tesfaldet2017two}, we perform a human perceptual study to evaluate the perceived realism of the synthesized examples. We randomly select 20 different human users. Each user is sequentially presented a pair of synthesized and real dynamic textures in a random order, and asked to select which one is fake after viewing them for a specified exposure time. The ``fooling'' rate, which is the user error rate in discriminating real versus synthesized dynamic textures, is calculated to measure the realism of the synthesized results. Higher ``fooling'' rate indicates more realistic and convincing synthesized dynamic textures. ``Perfect'' synthesized results corresponds to a fooling rate of 50\%  (i.e., random guess), meaning that the users are unable to distinguish between the synthesized and real examples. The number of pairwise comparisons presented to each user is 36 (12 categories $\times$ 3 examples). The exposure time is chosen from discrete durations between 0.3 and 3.6 seconds. 

We compare our model with three baseline methods, such as LDS (linear dynamic system)  \cite{doretto2003dynamic}, TwoStream \cite{tesfaldet2017two} and MoCoGAN \cite{tulyakov2017mocogan}, for dynamic texture synthesis in terms of ``fooling'' rate on 12 dynamic texture videos (e.g., waterfall, burning fire, waving flag, etc). 

LDS represents dynamic textures by a linear autoregressive model; TwoStream method synthesizes dynamic textures by matching the feature statistics extracted from two pre-trained convolutional networks between synthesized and observed examples; and MoCoGAN  is a motion and content decomposed generative adversarial network for video generation. 

\begin{figure}
\centering	
\includegraphics[width=.94\linewidth, height=.53\linewidth]{./
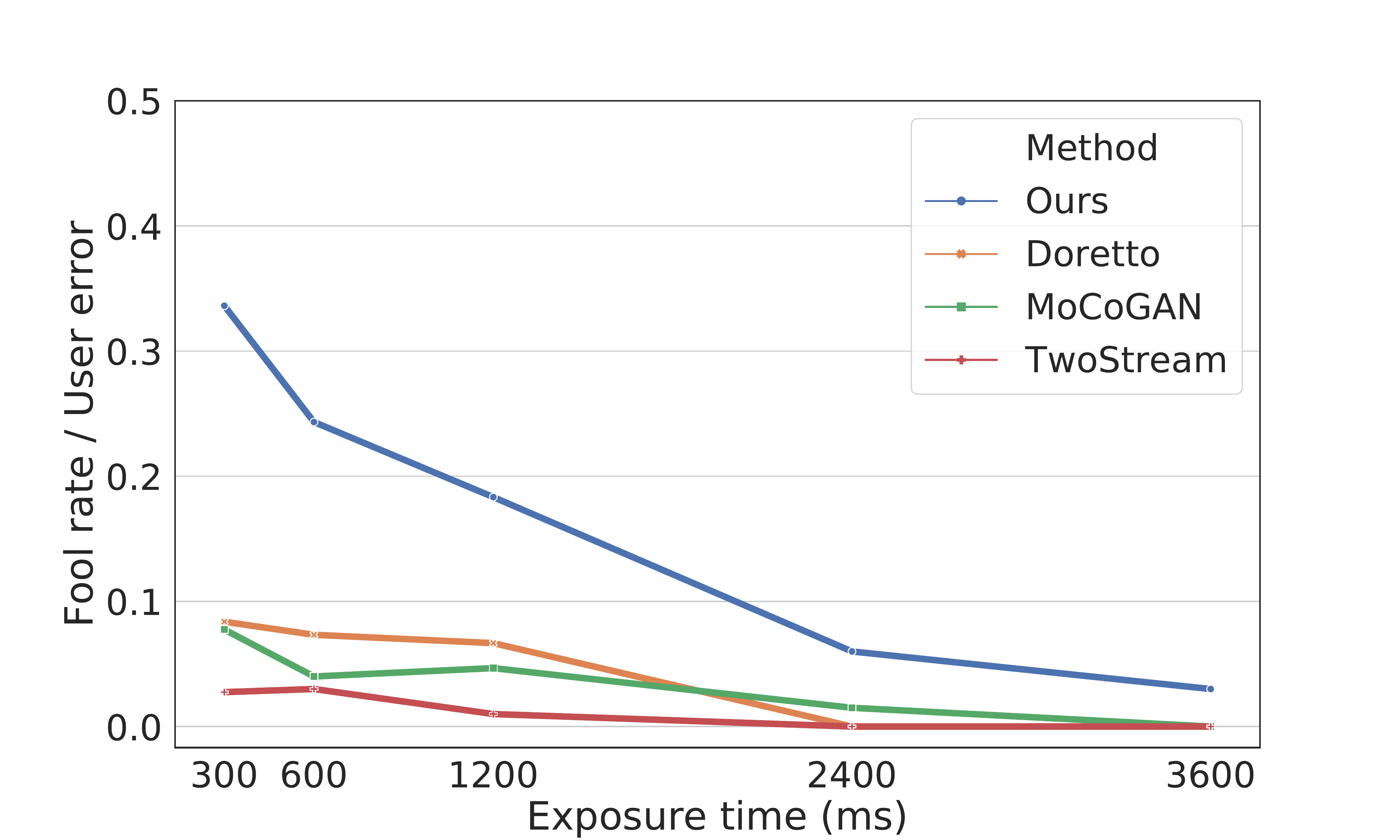}
\caption{Limited time pairwise comparison results. Each curve shows the ``fooling'' rates (realism) over different exposure times.}
\label{fig:humanStudy}
\end{figure}

Figure \ref{fig:humanStudy} summarizes the comparative result by showing the ``fooling'' rate as a function of exposure time across methods. We can find that as the given exposure time becomes longer, it becomes easier for the users to observe the difference between the real and synthesized dynamic textures. More specifically, the ``fooling'' rate decreases as exposure time increases, and then remains at the same level for longer exposures. Overall, our method can generate more realistic dynamic textures than other baseline methods. The result also shows that the linear model (i.e., LDS) outperforms the more sophisticated baselines (i.e., TwoStream and MoCoGAN). The reason is because when learning from a single example, the MoCoGAN may not fit the training data very well due to the unstable and complicated adversarial training scheme as well as a large number of parameters to be learned, and the TwoStream method has a limitation that it cannot handle dynamic textures that have structured background (e.g., burning fire heating a pot).

\subsection{Experiment 2: Learn to generate action patterns with appearance consistency}
\label{sec:action}
We learn the model from multiple examples with different appearances by using a 100-dimensional appearance vector. We infer the appearance vector and the initial state via a 15-step Langevin dynamics within each iteration of the learning process. We learn the model using the Weizmann action dataset \cite{gorelick2007actions}, which contains 81 videos of 9 people performing 9 actions, including jacking, jumping, walking, etc, as well as an animal action dataset that includes 20 videos of 10 animals performing running and walking collected from the Internet. Each video is scaled to $64 \times 64$ pixels $\times 30$ frames. We adopt the same structure of the model as the one in Section \ref{sec:dynamicTexture}, except that the emission model takes the concatenation of the appearance vector and the hidden state as input. For each experiment, a single model is trained on the whole dataset without annotations. The dimensions of the hidden state $s$ and the Gaussian noise $\xi$ are set to be 100 and 50 respectively for the Weizmann action dataset, and 3 and 100 for the animal action dataset.  

\begin{figure}[h]
\centering	
\rotatebox{90}{\hspace{4mm}{\footnotesize syn1}}	
\includegraphics[height=.14\linewidth, width=.14\linewidth]{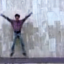} 
\includegraphics[height=.14\linewidth, width=.14\linewidth]{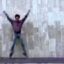} 
\includegraphics[height=.14\linewidth, width=.14\linewidth]{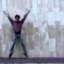} 
\includegraphics[height=.14\linewidth, width=.14\linewidth]{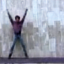} 
\includegraphics[height=.14\linewidth, width=.14\linewidth]{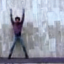}
\includegraphics[height=.14\linewidth, width=.14\linewidth]{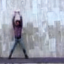}\\
\rotatebox{90}{\hspace{4mm}{\footnotesize syn2}}	
\includegraphics[height=.14\linewidth, width=.14\linewidth]{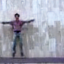} 
\includegraphics[height=.14\linewidth, width=.14\linewidth]{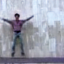} 
\includegraphics[height=.14\linewidth, width=.14\linewidth]{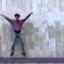} 
\includegraphics[height=.14\linewidth, width=.14\linewidth]{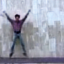} 
\includegraphics[height=.14\linewidth, width=.14\linewidth]{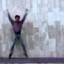}
\includegraphics[height=.14\linewidth, width=.14\linewidth]{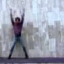} \\ person 1 \\
\vspace{2mm} 
\rotatebox{90}{\hspace{4mm}{\footnotesize syn1}}	
\includegraphics[height=.14\linewidth, width=.14\linewidth]{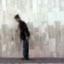} 
\includegraphics[height=.14\linewidth, width=.14\linewidth]{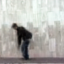}
\includegraphics[height=.14\linewidth, width=.14\linewidth]{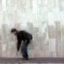} 
\includegraphics[height=.14\linewidth, width=.14\linewidth]{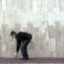} 
\includegraphics[height=.14\linewidth, width=.14\linewidth]{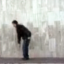} 
\includegraphics[height=.14\linewidth, width=.14\linewidth]{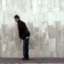} \\
\rotatebox{90}{\hspace{4mm}{\footnotesize syn2}}	
\includegraphics[height=.14\linewidth, width=.14\linewidth]{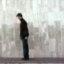} 
\includegraphics[height=.14\linewidth, width=.14\linewidth]{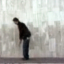}
\includegraphics[height=.14\linewidth, width=.14\linewidth]{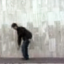} 
\includegraphics[height=.14\linewidth, width=.14\linewidth]{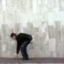} 
\includegraphics[height=.14\linewidth, width=.14\linewidth]{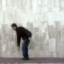} 
\includegraphics[height=.14\linewidth, width=.14\linewidth]{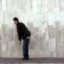} 
\\ person 2 \\

(a) synthesizing human actions (Weizmann dataset)\\
\vspace{1mm}
\rotatebox{90}{\hspace{4mm}{\footnotesize syn1}}	
\includegraphics[height=.14\linewidth, width=.14\linewidth]{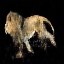} 
\includegraphics[height=.14\linewidth, width=.14\linewidth]{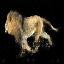} 
\includegraphics[height=.14\linewidth, width=.14\linewidth]{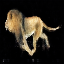}
\includegraphics[height=.14\linewidth, width=.14\linewidth]{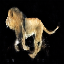} 
\includegraphics[height=.14\linewidth, width=.14\linewidth]{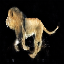}
\includegraphics[height=.14\linewidth, width=.14\linewidth]{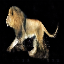} \\
\rotatebox{90}{\hspace{4mm}{\footnotesize syn2}}	
\includegraphics[height=.14\linewidth, width=.14\linewidth]{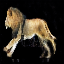} 
\includegraphics[height=.14\linewidth, width=.14\linewidth]{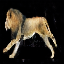}
\includegraphics[height=.14\linewidth, width=.14\linewidth]{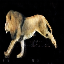} 
\includegraphics[height=.14\linewidth, width=.14\linewidth]{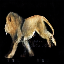} 
\includegraphics[height=.14\linewidth, width=.14\linewidth]{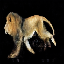}
\includegraphics[height=.14\linewidth, width=.14\linewidth]{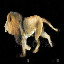}  \\ lion \\
\vspace{2mm}
\rotatebox{90}{\hspace{4mm}{\footnotesize syn1}}	
\includegraphics[height=.14\linewidth, width=.14\linewidth]{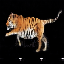} 
\includegraphics[height=.14\linewidth, width=.14\linewidth]{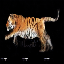} 
\includegraphics[height=.14\linewidth, width=.14\linewidth]{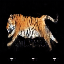} 
\includegraphics[height=.14\linewidth, width=.14\linewidth]{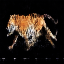} 
\includegraphics[height=.14\linewidth, width=.14\linewidth]{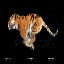}
\includegraphics[height=.14\linewidth, width=.14\linewidth]{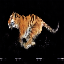}  \\
\rotatebox{90}{\hspace{4mm}{\footnotesize syn2}}	
\includegraphics[height=.14\linewidth, width=.14\linewidth]{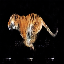} 
\includegraphics[height=.14\linewidth, width=.14\linewidth]{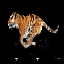} 
\includegraphics[height=.14\linewidth, width=.14\linewidth]{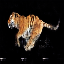} 
\includegraphics[height=.14\linewidth, width=.14\linewidth]{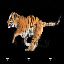} 
\includegraphics[height=.14\linewidth, width=.14\linewidth]{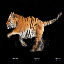} 
\includegraphics[height=.14\linewidth, width=.14\linewidth]{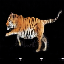} \\ tiger \\
(b) synthesizing animal actions (animal action dataset)\\
\vspace{1mm}
   \caption{Generated action patterns. For each inferred appearance vector, two synthesized videos are displayed.}	
\label{fig:motion}
\end{figure}

Figure \ref{fig:motion} shows some synthesized results for each experiment. To synthesize video, we randomly pick an appearance vector inferred from the observed video and generate new motion pattern for that specified appearance vector by the learned model with a noise sequence of $\{\xi_t, t=1,..,T\}$ and an initial state $s_0$ sampled from Gaussian white noise. We show two different synthesized motions for each appearance vector. With a fixed appearance, the learned model can generate diverse motions with consistent appearance.


  
\begin{figure}[h]
\centering	
\includegraphics[height=.11\linewidth]{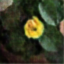}
\includegraphics[height=.11\linewidth]{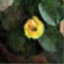}
\includegraphics[height=.11\linewidth]{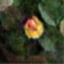}
\includegraphics[height=.11\linewidth]{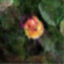}
\includegraphics[height=.11\linewidth]{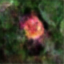}
\includegraphics[height=.11\linewidth]{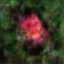}
\includegraphics[height=.11\linewidth]{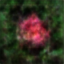}
\includegraphics[height=.11\linewidth]{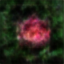}\\ \vspace{0.5mm}

\includegraphics[height=.11\linewidth]{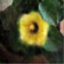}
\includegraphics[height=.11\linewidth]{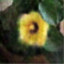}
\includegraphics[height=.11\linewidth]{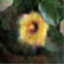}
\includegraphics[height=.11\linewidth]{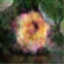}
\includegraphics[height=.11\linewidth]{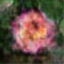}
\includegraphics[height=.11\linewidth]{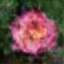}
\includegraphics[height=.11\linewidth]{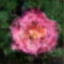}
\includegraphics[height=.11\linewidth]{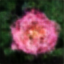}\\ \vspace{0.5mm}

\includegraphics[height=.11\linewidth]{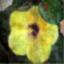}
\includegraphics[height=.11\linewidth]{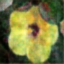}
\includegraphics[height=.11\linewidth]{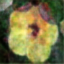}
\includegraphics[height=.11\linewidth]{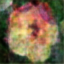}
\includegraphics[height=.11\linewidth]{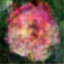}
\includegraphics[height=.11\linewidth]{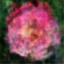}
\includegraphics[height=.11\linewidth]{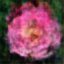}
\includegraphics[height=.11\linewidth]{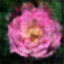} \\
(a) blooming  \\ \vspace{1mm}
\includegraphics[height=.11\linewidth]{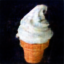}
\includegraphics[height=.11\linewidth]{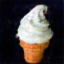}
\includegraphics[height=.11\linewidth]{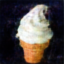}
\includegraphics[height=.11\linewidth]{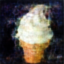}
\includegraphics[height=.11\linewidth]{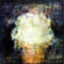}
\includegraphics[height=.11\linewidth]{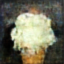}
\includegraphics[height=.11\linewidth]{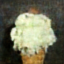}
\includegraphics[height=.11\linewidth]{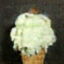}\\ \vspace{0.5mm}

\includegraphics[height=.11\linewidth]{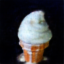}
\includegraphics[height=.11\linewidth]{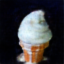}
\includegraphics[height=.11\linewidth]{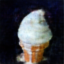}
\includegraphics[height=.11\linewidth]{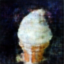}
\includegraphics[height=.11\linewidth]{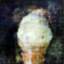}
\includegraphics[height=.11\linewidth]{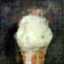}
\includegraphics[height=.11\linewidth]{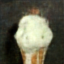}
\includegraphics[height=.11\linewidth]{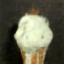} \\ \vspace{0.5mm}

\includegraphics[height=.11\linewidth]{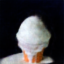}
\includegraphics[height=.11\linewidth]{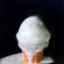}
\includegraphics[height=.11\linewidth]{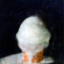}
\includegraphics[height=.11\linewidth]{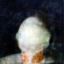}
\includegraphics[height=.11\linewidth]{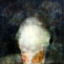}
\includegraphics[height=.11\linewidth]{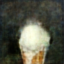}
\includegraphics[height=.11\linewidth]{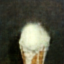}
\includegraphics[height=.11\linewidth]{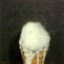}\\
(b) melting \\ 
\caption{Video interpolation by interpolating between appearance latent vectors of videos at the two ends. For each example, each column is one synthesized video. We show 3 frames for each video in each column. }	
\label{fig:Interpolation}
\end{figure}

Figure \ref{fig:Interpolation} shows two examples of video interpolation by interpolating between appearance vectors of videos at the two ends. We conduct these  experiments on some videos selected from categories ``blooming'' and ``melting'' in the dataset of \cite{zhou2016learning}. For each example, the videos at the two ends are generated with the appearance vectors inferred from two observed videos. Each video in the middle is obtained by first interpolating the appearance vectors of the two end videos, and then generating the videos using the dynamic generator. All the generated videos use the same set of noise sequence $\{\xi_t\}$ and $s_0$ randomly sampled from Gaussian white noise. We observe smooth transitions in contents and motions of all the generated videos and that the intermediate videos are also physically plausible.

We compare with MoCoGAN and TGAN \cite{saito2017temporal} by training on 9 selected categories (e.g., PlayingCello, PlayingDaf, PlayingDhol, PlayingFlute, PlayingGuitar, PlayingPiano, PlayingSitar, PlayingTabla, and PlayingViolin) of videos in the UCF101 \cite{soomro2012ucf101} database and following \cite{saito2017temporal} to compute the inception score. Table \ref{ic_score} shows comparison results. Our model outperforms the MoCoGAN and TGAN in terms of inception score. 

\begin{table}[h]
\centering
\caption{Inception score for models trained on 9 classes of videos in UCF101 database.}
\label{tab:recovery}
\begin{tabular}{|c|ccc|}
\hline 
Reference    & ours  &  MoCoGAN  &TGAN   \\ \hline \hline
11.05$\pm$0.16  & {\bf 8.21$\pm$0.09} & 4.40$\pm$0.04  & 5.48$\pm$0.06 \\ 
\hline
\end{tabular}
\label{ic_score}
\end{table}

\begin{figure}[h]
\centering	
\includegraphics[height=.17\linewidth, width=.17\linewidth]{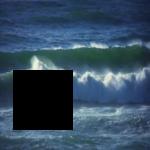} 
\includegraphics[height=.17\linewidth, width=.17\linewidth]{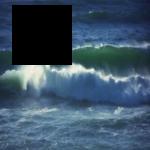} 
\includegraphics[height=.17\linewidth, width=.17\linewidth]{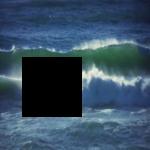} 
\includegraphics[height=.17\linewidth, width=.17\linewidth]{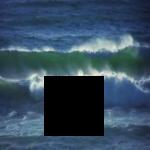} 
\includegraphics[height=.17\linewidth, width=.17\linewidth]{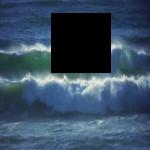} \\
\vspace{0.5mm}
\includegraphics[height=.17\linewidth, width=.17\linewidth]{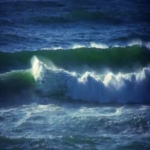} 
\includegraphics[height=.17\linewidth, width=.17\linewidth]{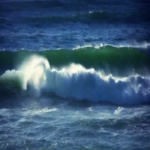} 
\includegraphics[height=.17\linewidth, width=.17\linewidth]{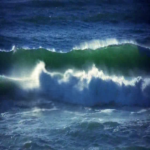} 
\includegraphics[height=.17\linewidth, width=.17\linewidth]{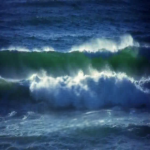} 
\includegraphics[height=.17\linewidth, width=.17\linewidth]{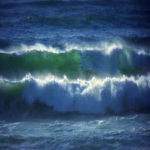} \\
(a) ocean\\  
\vspace{0.5mm}
\includegraphics[height=.17\linewidth, width=.17\linewidth]{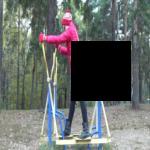} 
\includegraphics[height=.17\linewidth, width=.17\linewidth]{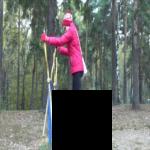} 
\includegraphics[height=.17\linewidth, width=.17\linewidth]{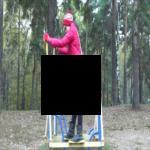} 
\includegraphics[height=.17\linewidth, width=.17\linewidth]{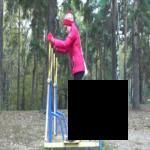} 
\includegraphics[height=.17\linewidth, width=.17\linewidth]{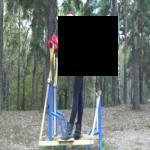} \\
\vspace{0.5mm}
\includegraphics[height=.17\linewidth, width=.17\linewidth]{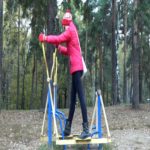} 
\includegraphics[height=.17\linewidth, width=.17\linewidth]{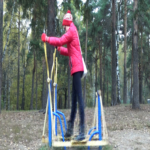} 
\includegraphics[height=.17\linewidth, width=.17\linewidth]{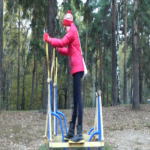} 
\includegraphics[height=.17\linewidth, width=.17\linewidth]{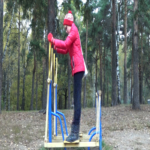} 
\includegraphics[height=.17\linewidth, width=.17\linewidth]{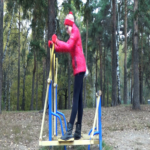} \\
(b) playing\\
 \vspace{0.5mm}
\includegraphics[height=.184\linewidth]{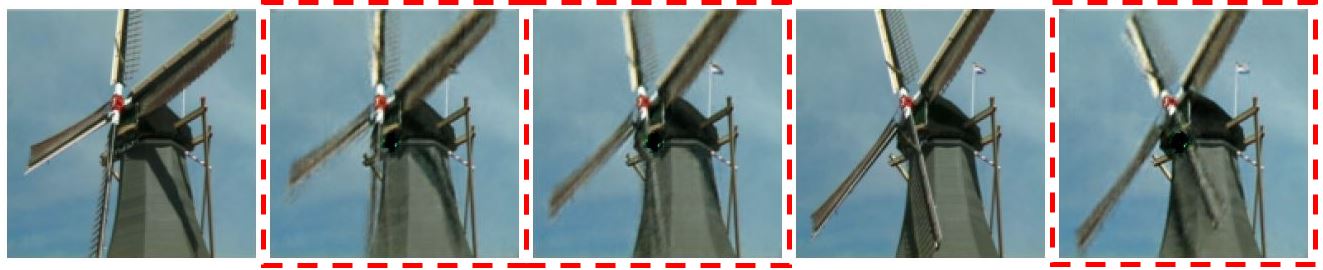} \\
(c) windmill\\
 \vspace{0.5mm}
\includegraphics[height=.186\linewidth]{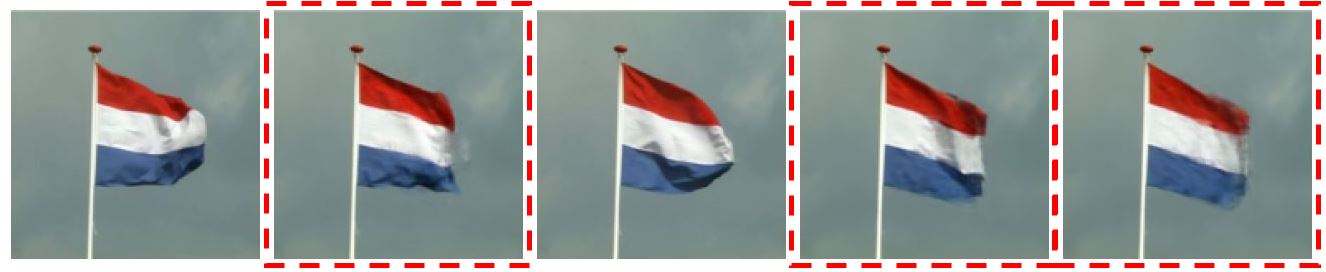} \\
(d) flag\\ 
\caption{Learning from occluded videos. (a,b) For each experiment, the first row displays a segment of the occluded sequence with black masks. The second row shows the corresponding segment of the recovered sequence. (c,d) The 3 frames with red bounding box are recovered by the learning algorithm, and they are occluded in the training stage. Each video has 70 frames and 50$\%$ frames are randomly occluded. 
 }	
\label{fig:recovery}
\end{figure}

\subsection{Experiment 3: Learn from incomplete data}
\label{sec:recovery}
Our model can learn from videos with occluded pixels and frames. We adapt our algorithm to this task with minimal modification involving the computation
of $\sum_{t=1}^{T}\|x_t - G_\beta(s_t)\|^2 $. In the setting of learning from fully observed videos, it is computed by summing over all the pixels of the video frames, while in the setting of learning from partially visible videos, we compute it by summing over only the visible pixels of the video frames. Then we can continue to use the alternating back-propagation through time (ABPTT) algorithm to infer $\{\xi_t, t=1,...,T\}$ and $s_0$, and then learn $\beta$ and $\alpha$. With inferred $\{\xi_t\}$ and $s_0$, and learned $\beta$ and $\alpha$, the video with occluded pixels or frames can be automatically recovered by $G_{\beta}(s_t)$, where the hidden state can be recursively computed by $s_{t}=F_{\alpha}(s_{t-1}, \xi_t)$. 

Eventually, our model can achieve the
following tasks: (1) recover the occluded pixels of training videos. (2) synthesize new videos by the learned model. (3) recover the occluded pixels of testing videos using the learned model. Different from those inpainting methods where the prior model has already been given or learned from fully observed training data, our recovery experiment is about an unsupervised learning task, where the ground truths of the occluded pixels are unknown in training the model for recovery. It is also worth mentioning that learning from incomplete data can be difficult for GANs (e.g., MoCoGAN), because of their lack of an adaptive inference process in the training stage. Here ``adaptive'' means the inference can be performed on input images with different sets of occluded pixels. 

We test our recovery algorithm on 6 video sequences collected from DynTex++ dataset. Each input video is of the size 150 pixels $\times$ 150 pixels $\times$ 70 frames. The emission model is a top-down deconvolutional neural network that maps a 100-dimensional state vector $s_t$ to the image frame of size $150 \times 150 \times 70$ by 7 layers of deconvolutions with numbers of channels $\{512, 512, 256, 128, 64, 64, 3\}$, kernel sizes $\{4, 4, 4, 4, 4, 4, 7\}$, and up-sampling factors $\{2, 2, 2, 2, 3, 3, 1\}$ at different layers from top to  bottom. We use the same transition model and the same parameter setting as in Section \ref{sec:dynamicTexture}, except that the standard deviation of residual error is $\sigma=0.5$. We run 7,000 iterations to recover each video. The length of chunk is 70.  
  
  We have two types of occlusions: 
(1) single region mask occlusion, where a $60 \times 60$ mask is randomly placed on each $150 \times 150$ image frame of each video. (2) missing image frames, where 50$\%$ of the image frames are randomly blocked in each video. For each type of occlusion experiment, we measure the recovery errors by the average per pixel difference between the recovered video sequences and the original ones (The
range of pixel intensities is [0, 255]), and compare with STGCN \cite{xie2017synthesizing}, which is a spatial-temporal deep convolutional energy-based model that can recover missing pixels of videos by synthesis during the learning process. 
We also report results obtained by generic spatial-temporal Markov random field models with potentials that are $\ell_1$ or $\ell_2$ difference between pixels of nearest neighbors that are defined in both spatial and temporal domains, and the recovery is accomplished by synthesizing missing pixels via Gibbs sampling. Table \ref{recoveryExp} shows the comparison results. Some qualitative results for recovery by our models are displayed in Figure \ref{fig:recovery}. 

\begin{table}[h]
\caption{Recovery errors in occlusion experiments}\label{recoveryExp}
\vskip 0.02in
\begin{center}
\begin{footnotesize}
(a) single region masks \\
\begin{tabular}{|c|c|c|c|c|}
\hline
           & ours  & STGCN  & MRF-$\ell_1$      & MRF-$\ell_2$  \\ \hline \hline
flag    & \textbf{7.8782}  & 8.1636  & 10.6586  & 12.5300\\ \hline
fountain & \textbf{5.6988} & 6.0323  & 11.8299  & 12.1696\\ \hline
ocean  &  \textbf{3.3966}  & 3.4842  & 8.7498 & 9.8078\\ \hline
playing  & \textbf{4.9251} & 6.1575  & 15.6296& 15.7085\\ \hline
sea world & \textbf{5.6596} & 5.8850 & 12.0297  & 12.2868\\ \hline
windmill & \textbf{6.6827} & 7.8858 & 11.7355   & 13.2036\\ \hline 
Avg.   & \textbf{5.7068}&  6.2681 &  11.7722   & 12.6177\\ \hline
\end{tabular}
\vskip 0.08in
(b) 50$\%$ missing frames\\
\begin{tabular}{|c|c|c|c|c|}
\hline
         & ours  &  STGCN  & MRF-$\ell_1$      & MRF-$\ell_2$  \\ \hline \hline
flag    &  \textbf{5.0874} & 5.5992 & 10.7171  & 12.6317\\ \hline
fountain & \textbf{5.5669} & 8.0531 & 19.4331  & 13.2251\\ \hline
ocean  &  \textbf{3.3666}  & 4.0428  & 9.0838 & 9.8913\\ \hline
playing  & \textbf{5.2563} & 7.6103 & 22.2827& 17.5692\\ \hline
sea world & \textbf{4.0682} & 5.4348 & 13.5101  &12.9305\\ \hline
windmill & \textbf{6.9267} & 7.5346 & 13.3364  & 12.9911 \\ \hline 
Avg.  & \textbf{5.0454} &  6.3791 &  14.7272   & 13.2065\\ \hline
\end{tabular}
\end{footnotesize}
\end{center}
\end{table}

\subsection{Experiment 4: Learn to remove content}

The dynamic generarator model can be used to remove undesirable content in the video for background inpainting. The basic idea is as follows. We first manually mask the undesirable moving object in each frame of the video, and then learn the model from the masked video with the recovery algorithm that we used in Section \ref{sec:recovery}. Since there are neither clues in the masked video nor prior knowledge to infer the occluded object, it turns out to be that the recovery algorithm will inpaint the empty region with the background.  

\begin{figure}[t]
\centering	
\includegraphics[height=.18\linewidth, width=.18\linewidth]{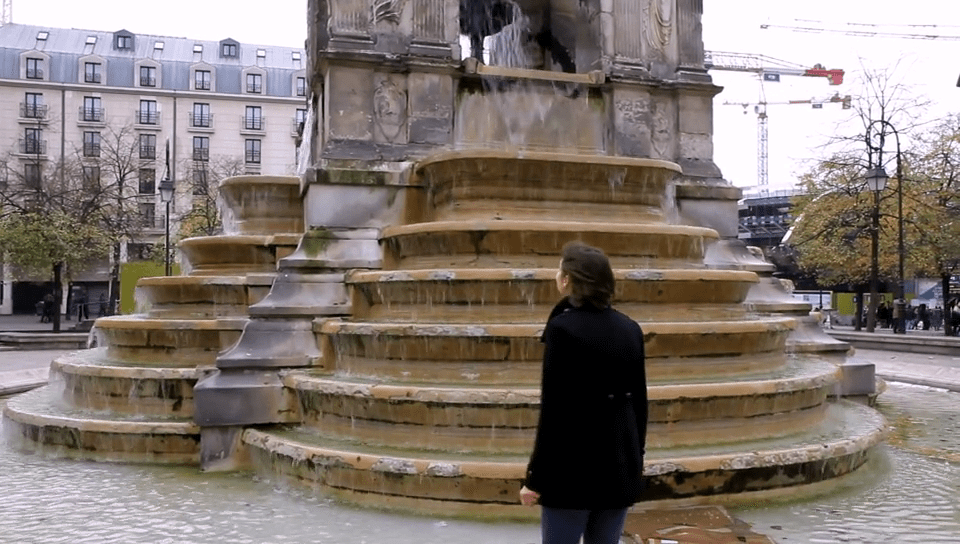}
\includegraphics[height=.18\linewidth, width=.18\linewidth]{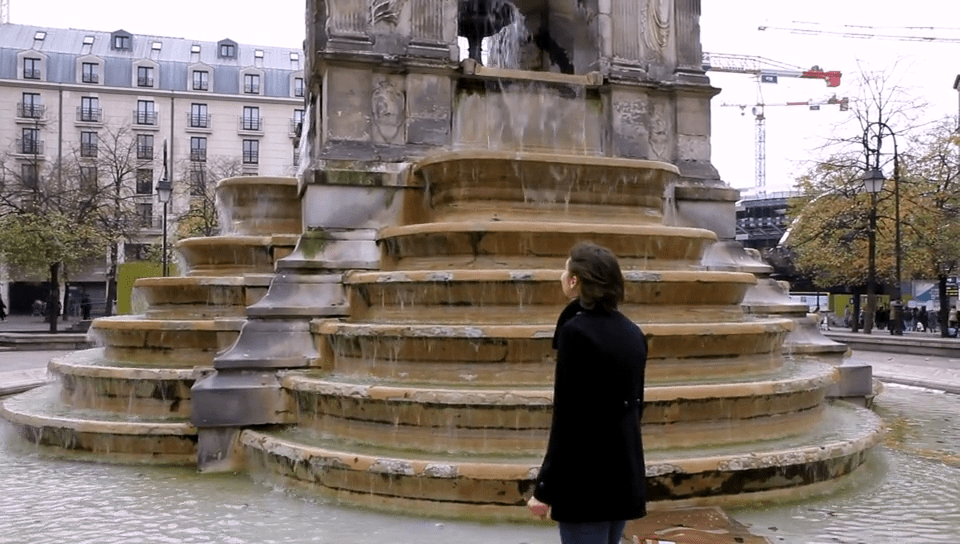}
\includegraphics[height=.18\linewidth, width=.18\linewidth]{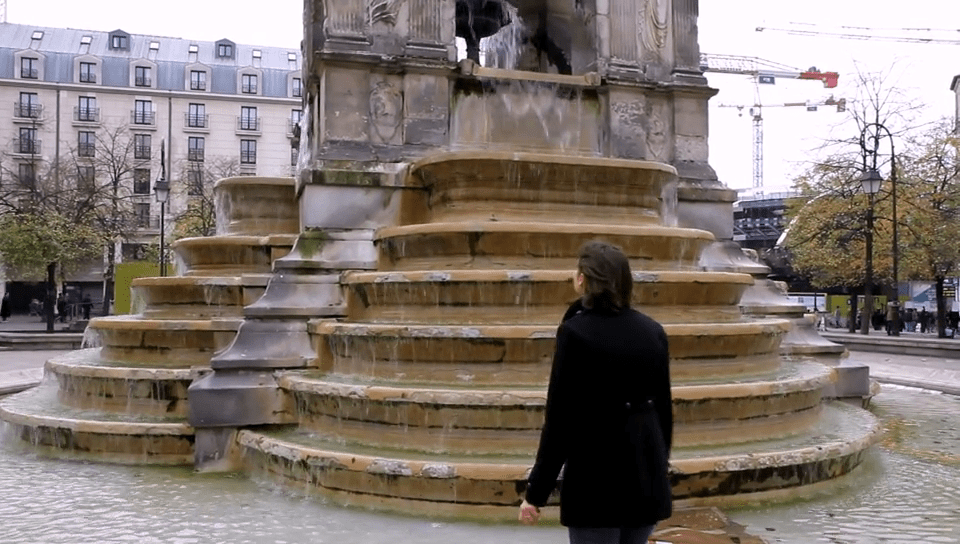}
\includegraphics[height=.18\linewidth, width=.18\linewidth]{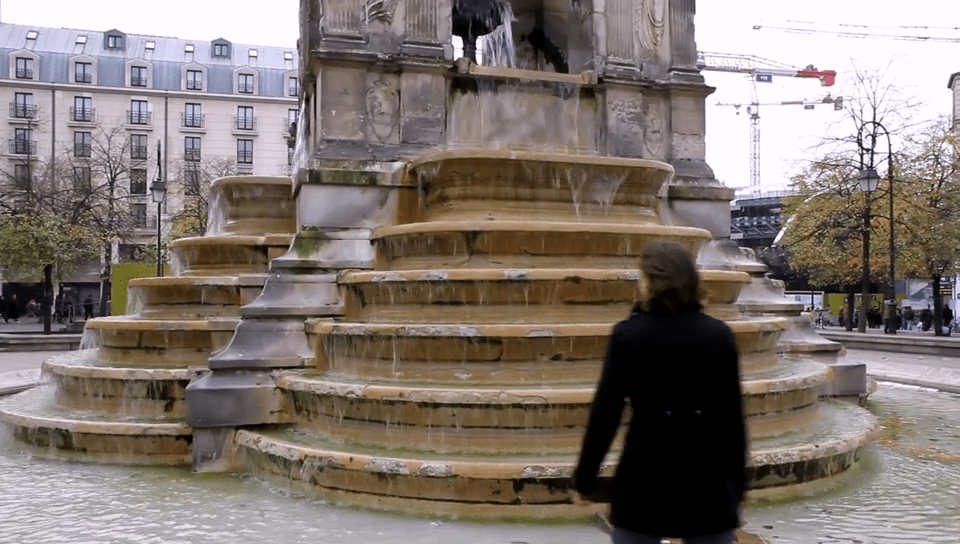}	
\includegraphics[height=.18\linewidth, width=.18\linewidth]{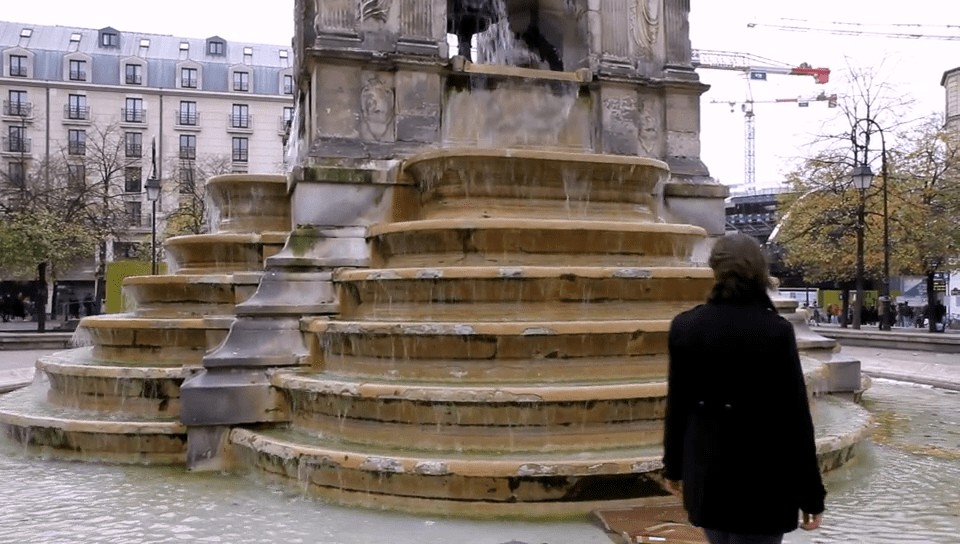} \\ \vspace{1mm}
\includegraphics[height=.18\linewidth, width=.18\linewidth]{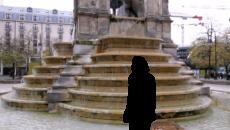} 	
\includegraphics[height=.18\linewidth, width=.18\linewidth]{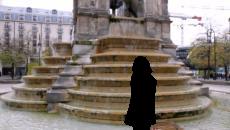} 	
\includegraphics[height=.18\linewidth, width=.18\linewidth]{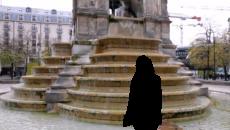} 	
\includegraphics[height=.18\linewidth, width=.18\linewidth]{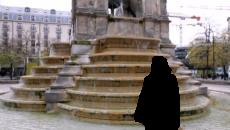} 	
\includegraphics[height=.18\linewidth, width=.18\linewidth]{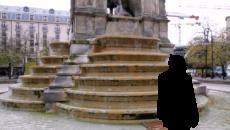} 
\\ \vspace{1mm}
\includegraphics[height=.18\linewidth, width=.18\linewidth]{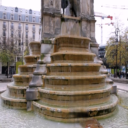}
\includegraphics[height=.18\linewidth, width=.18\linewidth]{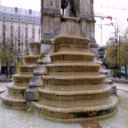}
\includegraphics[height=.18\linewidth, width=.18\linewidth]{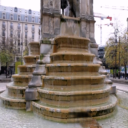}
\includegraphics[height=.18\linewidth, width=.18\linewidth]{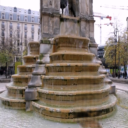}	
\includegraphics[height=.18\linewidth, width=.18\linewidth]{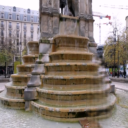} 
\\ (a)  removing a walking person in front of fountain
\\ \vspace{2mm}
\includegraphics[height=.18\linewidth, width=.18\linewidth]{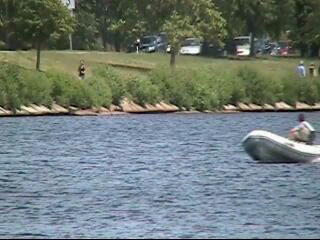} 
\includegraphics[height=.18\linewidth, width=.18\linewidth]{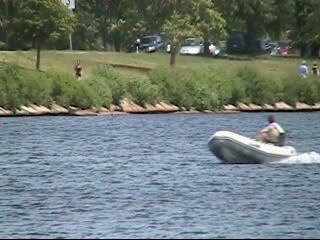}
\includegraphics[height=.18\linewidth, width=.18\linewidth]{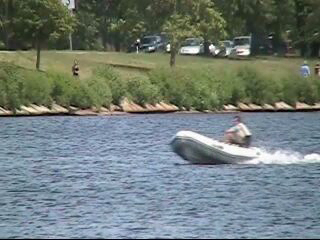}
\includegraphics[height=.18\linewidth, width=.18\linewidth]{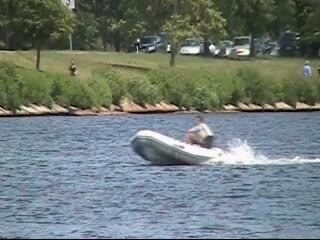}	
\includegraphics[height=.18\linewidth, width=.18\linewidth]{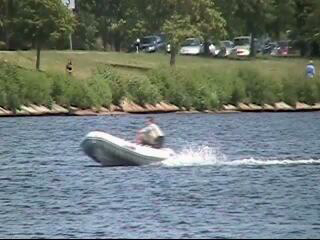}
\\ \vspace{1mm}
\includegraphics[height=.18\linewidth, width=.18\linewidth]{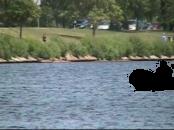} 	
\includegraphics[height=.18\linewidth, width=.18\linewidth]{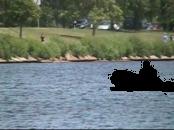} 
\includegraphics[height=.18\linewidth, width=.18\linewidth]{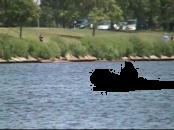} 
\includegraphics[height=.18\linewidth, width=.18\linewidth]{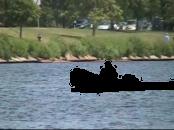} 
\includegraphics[height=.18\linewidth, width=.18\linewidth]{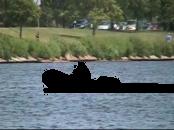} 
\\ \vspace{1mm}
\includegraphics[height=.18\linewidth, width=.18\linewidth]{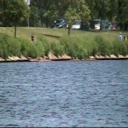} 
\includegraphics[height=.18\linewidth, width=.18\linewidth]{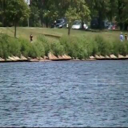} 
\includegraphics[height=.18\linewidth, width=.18\linewidth]{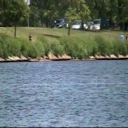} 
\includegraphics[height=.18\linewidth, width=.18\linewidth]{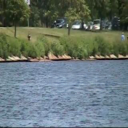} 	
\includegraphics[height=.18\linewidth, width=.18\linewidth]{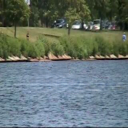}\\
 (b) removing a moving boat in the lake
   	\caption{Learn to remove content for background inpainting. For each experiment, the first row displays 5 image frames of the original video. The second row displays the corresponding image frames with black mask occluding the target to be removed. The third row shows the inpainting results by our method. (a) walking person. (b) moving boat.}   	
\label{fig:background_inpainiting}
\end{figure}

 Figure \ref{fig:background_inpainiting} shows two examples of removals of (a) a walking person and (b) a moving boat respectively. The videos are collected from \cite{braham2016deep}. For each example, the first row displays 5 frames of the original video. The second row shows the corresponding frames with masks occluding the target to be removed. The third row presents the inpainting results by our algorithm. The video size is $128 \times 128 \times 104$ in example (a) and $128 \times 128 \times 150$ in example (b). We adopt the same transition model as the one in Section \ref{sec:recovery}, and an emission model that has 7 layers of deconvolutions with kernel size of 4, up-sampling factor of 2, and numbers of channels $\{512, 512, 512, 256, 128, 64, 3\}$ at different layers from top to bottom. The emission model maps the 100-dimensional state vector to the image frame of size $128 \times 128$ pixels.
 
  The experiment is different from the background inpainting by \cite{xie2017synthesizing}, where the empty regions of the video are inpainted by directly sampling from a probability distribution of pixels in empty region conditioned on visible pixels. As to our model, we inpaint the empty regions of the video by inferring all the latent variables by Langevin dynamics. 

\subsection{Experiment 5: Learn to animate static image}
A conditional version of the dynamic generator model can be used for video prediction given a static image. Specifically, we learn a mapping from a static image frame to the subsequent frames. We incorporate an extra encoder $E_{\gamma}$, where $\gamma$ denotes the weight and bias parameters of the encoder, to map the first image frame $x^{(i)}_0$ into its appearance or content vector $a^{(i)}$ and state vector $s_0^{(i)}$. The dynamic generator takes the state vector $s_0^{(i)}$ as the initial state and uses the appearance vector $a^{(i)}$ to generate the subsequent video frames $\{x_t^{(i)}, t=1,...,T\}$ for the $i$-th video. The conditional model is of the following form
\begin{eqnarray}
 && [s_{0}^{(i)}, a^{(i)}]= E_\gamma(x_{0}^{(i)}), \\
 && s_{t}^{(i)} = F_\alpha(s_{t-1}^{(i)}, \xi_t^{(i)}),  \label{eq:prediction:t} \\
 && x_t^{(i)} = G_\beta(s_t^{(i)}, a^{(i)}) + \epsilon_t^{(i)}. \label{eq:prediction:e}
\end{eqnarray}
We learn both the encoder and the dynamic generator (i.e., transition model and emission model) together by alternating back-propagation through time. The appearance vector and the initial state are no longer hidden variables that need to be inferred in training. Once the model is learned, given a testing static image, the learned encoder $E_{\gamma}$ extracts from it  the appearance vector and the initial state vector, which generate a sequence of images by the dynamic generator.

\begin{figure}
\centering	

\includegraphics[height=.13\linewidth, width=.13\linewidth]{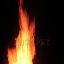} \hspace{1mm}
\includegraphics[height=.13\linewidth, width=.13\linewidth]{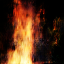}
\includegraphics[height=.13\linewidth, width=.13\linewidth]{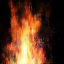}
\includegraphics[height=.13\linewidth, width=.13\linewidth]{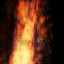}
\includegraphics[height=.13\linewidth, width=.13\linewidth]{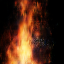}
\includegraphics[height=.13\linewidth, width=.13\linewidth]{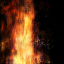}
\includegraphics[height=.13\linewidth, width=.13\linewidth]{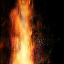} \\ \vspace{1mm}

\includegraphics[height=.13\linewidth, width=.13\linewidth]{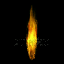} \hspace{1mm}
\includegraphics[height=.13\linewidth, width=.13\linewidth]{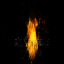}
\includegraphics[height=.13\linewidth, width=.13\linewidth]{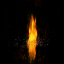}
\includegraphics[height=.13\linewidth, width=.13\linewidth]{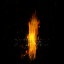}
\includegraphics[height=.13\linewidth, width=.13\linewidth]{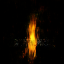}
\includegraphics[height=.13\linewidth, width=.13\linewidth]{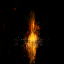}
\includegraphics[height=.13\linewidth, width=.13\linewidth]{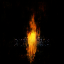} \\ \vspace{1mm}

\includegraphics[height=.13\linewidth, width=.13\linewidth]{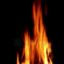} \hspace{1mm}
\includegraphics[height=.13\linewidth, width=.13\linewidth]{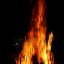}
\includegraphics[height=.13\linewidth, width=.13\linewidth]{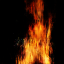}
\includegraphics[height=.13\linewidth, width=.13\linewidth]{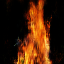}
\includegraphics[height=.13\linewidth, width=.13\linewidth]{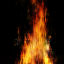}
\includegraphics[height=.13\linewidth, width=.13\linewidth]{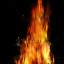}
\includegraphics[height=.13\linewidth, width=.13\linewidth]{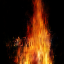} \\ 
(a) burning fire \\ \vspace{1mm} 
\includegraphics[height=.13\linewidth, width=.13\linewidth]{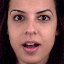} \hspace{1mm}
\includegraphics[height=.13\linewidth, width=.13\linewidth]{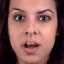}
\includegraphics[height=.13\linewidth, width=.13\linewidth]{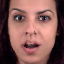}
\includegraphics[height=.13\linewidth, width=.13\linewidth]{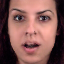}
\includegraphics[height=.13\linewidth, width=.13\linewidth]{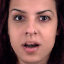}
\includegraphics[height=.13\linewidth, width=.13\linewidth]{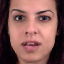}
\includegraphics[height=.13\linewidth, width=.13\linewidth]{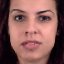} \\ 
\vspace{0.5mm}
\includegraphics[height=.13\linewidth, width=.13\linewidth]{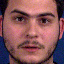} \hspace{1mm}
\includegraphics[height=.13\linewidth, width=.13\linewidth]{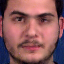}
\includegraphics[height=.13\linewidth, width=.13\linewidth]{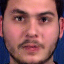}
\includegraphics[height=.13\linewidth, width=.13\linewidth]{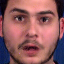}
\includegraphics[height=.13\linewidth, width=.13\linewidth]{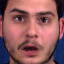}
\includegraphics[height=.13\linewidth, width=.13\linewidth]{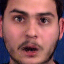}
\includegraphics[height=.13\linewidth, width=.13\linewidth]{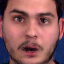}\\
\vspace{0.5mm}
\includegraphics[height=.13\linewidth, width=.13\linewidth]{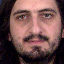} \hspace{1mm}
\includegraphics[height=.13\linewidth, width=.13\linewidth]{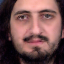}
\includegraphics[height=.13\linewidth, width=.13\linewidth]{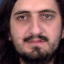}
\includegraphics[height=.13\linewidth, width=.13\linewidth]{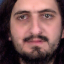}
\includegraphics[height=.13\linewidth, width=.13\linewidth]{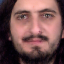}
\includegraphics[height=.13\linewidth, width=.13\linewidth]{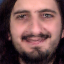}
\includegraphics[height=.13\linewidth, width=.13\linewidth]{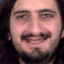}\\
(b) facial expression
   	\caption{Image-to-video prediction. For each example, the first image is the static image frame, and the rest are 6 frames of the predicted sequence.}	
\label{fig:prediction}
\end{figure}

We test our model on burning fire dataset \cite{xie2017synthesizing}, and MUG Facial Expression
dataset \cite{MUG2010}. The encoder has 3 convolutional layers with
numbers of channels $\{64, 128, 256\}$, filter sizes $\{5, 3, 3\}$ and sub-sampling factors $\{2, 2, 1\}$ at different layers, and one fully
connected layer with the output size equal to the dimension of the appearance vector (100) plus the dimension of the hidden state (80). The dimension of $\xi$ is 20. The other configurations are similar to what we used in Section \ref{sec:action}. We qualitatively display some results in Figure \ref{fig:prediction}, where each row is one example of image-to-video prediction. For each example, the left image is the static image frame for testing, and the rest are 6 frames of the predicted video sequence. The results show that the predicted frames by our method have fairly plausible motions. 

\section{Conclusion}

This paper studies a dynamic generator model for spatial-temporal processes. The model is a non-linear generalization of the linear state space model where the non-linear transformations in the transition and emission models are parameterized by neural networks. The model can be conveniently and efficiently learned by an alternating back-propagation through time (ABPTT) algorithm that alternatively samples from the posterior distribution of the latent noise vectors and then updates the model parameters. The model can be generalized by including random vectors to account for various sources of variations, and the learning algorithm can still apply to the generalized models. 

\section*{Project page} 

{\small The code and more results can be found at 
\url{http://www.stat.ucla.edu/~jxie/DynamicGenerator/DynamicGenerator.html}}

\section*{Acknowledgement}

{\small Part of the work was done while R. G. was an intern at Hikvision Research Institute during the summer of 2018. She thanks Director Jane Chen for her help and guidance. 

The work is supported by DARPA XAI project N66001-17-2-4029; ARO project W911NF1810296; ONR MURI project N00014-16-1-2007; and a Hikvision gift to UCLA.

We gratefully acknowledge the support of NVIDIA Corporation with the donation of the Titan Xp GPU used for this research.}
\medskip
\small
\bibliographystyle{aaai}
\bibliography{mybibfile}

\end{document}